\documentclass{article}

\usepackage{arxiv}

\usepackage[utf8]{inputenc} 
\usepackage[T1]{fontenc}    
\usepackage{microtype}      

\usepackage{graphicx}
\usepackage{xcolor}         

\usepackage{booktabs}       
\usepackage{multirow}       
\usepackage{multicol}       
\usepackage{colortbl}       

\usepackage{amsmath}  
\usepackage{amsfonts} 

\usepackage{url}            
\usepackage{nicefrac}       
\usepackage{lipsum}         
\usepackage{caption}        

\definecolor{cite-blue}{RGB}{30, 100, 170} 

\usepackage[numbers,sort&compress]{natbib}

\usepackage[colorlinks=true,citecolor=cite-blue,linkcolor=blue]{hyperref}

\usepackage{doi}

\title{RS-Gen: A Multi-Stage Agentic Framework for Reasoning and Search-Augmented Image Generation}

\author{
    Feifei Bian, Zhimin Zheng, Wei Deng, Daiguo Zhou and Jian Luan \\[2ex] 
    MiLM Plus, Xiaomi Inc. \\
    \texttt{\{bianfeifei,zhengzhimin,dengwei1,zhoudaiguo,luanjian\}@xiaomi.com} \\[4ex] 
}

\renewcommand{\headeright}{}
\renewcommand{\shorttitle}{\textbf{RS-Gen: A Multi-Stage Agentic Framework for Reasoning and Search-Augmented Image Generation}}

\hypersetup{
pdftitle={A template for the arxiv style},
pdfsubject={q-bio.NC, q-bio.QM},
pdfauthor={David S.~Hippocampus, Elias D.~Striatum},
pdfkeywords={First keyword, Second keyword, More},
}

\begin{document}
\maketitle

\begin{center}
    \centering
    \includegraphics[width=1.0\linewidth]{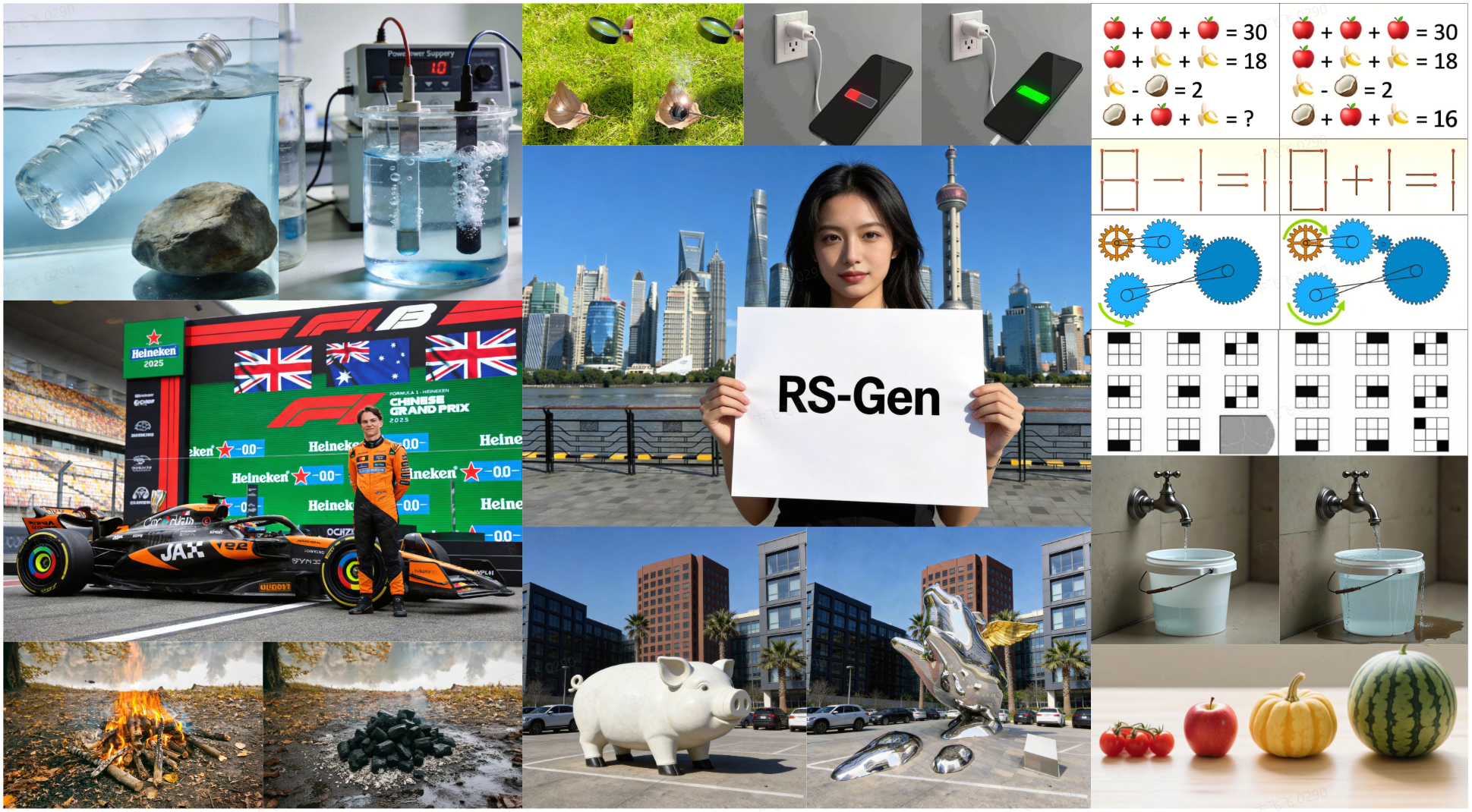}
    \captionof{figure}{Representative generation results by our proposed RS-Gen. By integrating external knowledge retrieval and logical reasoning mechanisms, RS-Gen achieves superior accuracy and fidelity in challenging tasks, including specific entity generation, logical puzzle-solving, visual reasoning, and physical evolution.} 
    \label{fig:show_case}
\end{center}
\vspace{1.5em} 

\begin{abstract}
Recent years have witnessed remarkable progress in image generation and editing, particularly regarding instruction following and visual fidelity. However, when handling ambiguous intentions, logical reasoning, and Out-of-Distribution (OOD) knowledge, existing image models often yield sub-optimal results due to a lack of deep reasoning capabilities and real-time external information. Although emerging unified understanding-and-generation models attempt to bridge this gap, they remain constrained by their intrinsic parameter scales and static knowledge gaps. Inspired by agentic paradigms, we propose \textbf{RS-Gen}: a plug-and-play, training-free, multi-stage image agentic framework. RS-Gen innovatively introduces a "Questioning-and-Solving" closed-loop mechanism to accurately identify logical issues and knowledge gaps, autonomously planning actions to bridge information deficits and execute deep logical reasoning. Extensive experiments demonstrate that RS-Gen significantly expands the capability boundaries of foundational image generation and editing models. Specifically, on the WISE\_Verified and RISEBench benchmarks, RS-Gen yields substantial absolute performance gains of 0.313 for Qwen-Image and 19.70 for Qwen-Image-Edit-2511, respectively, successfully elevating both to the state-of-the-art (SOTA) level among open-source models.
\end{abstract}


\section{Introduction}
In recent years, visual generation technologies, represented by diffusion models, have made breakthrough progress in the fields of image generation and editing. A series of cutting-edge models have emerged, such as FLUX.1-dev~\cite{flux2024}, Qwen-Image~\cite{wu2025qwenimagetechnicalreport}, Z-Image~\cite{team2025zimage}, and LongCat-Image~\cite{meituan2025longcatimage}, bringing the generated results to unprecedented heights in terms of high fidelity and realism. However, despite their ability to synthesize visually striking images, the trajectory toward truly "intelligent generation" remains impeded by three core bottlenecks. First, existing models exhibit an insufficient capability in parsing implicit intentions. They are heavily reliant on precise and explicit user prompts, lacking a deep semantic understanding of instructions that contain co-references or ambiguous expressions, which causes them to struggle with grounding and manipulating target objects during complex multi-turn interactive editing. Second, they lack visual logical reasoning capabilities. The current generation process is largely restricted to a superficial "text-to-pixel" statistical mapping. When confronted with complex visual tasks such as logical puzzle-solving or physical state evolution, models fail to comprehend the underlying causal logic and objective constraints, frequently yielding outputs that violate physical laws and common sense. Finally, these models suffer from inherent knowledge lag and the "hallucination" dilemma. Constrained by the static nature of training data and knowledge cut-offs, they are incapable of perceiving novel concepts or long-tail entities; when processing Out-of-Distribution (OOD) concepts, they are highly susceptible to factual errors and visual hallucinations, resulting in content that is severely decoupled from the real world.

To overcome these limitations, the academic community has engaged in active exploration. For instance, unified understanding-and-generation architectures, represented by Transfusion~\cite{zhou2024transfusion}, JanusFlow~\cite{ma2024janusflow}, and Bagel~\cite{deng2025bagel}, have made notable progress in intent comprehension and logical reasoning. However, they still exhibit significant bottlenecks when tackling multi-step reasoning or handling Out-of-Distribution (OOD) knowledge. Such limitations suggest that traditional monolithic architectural designs may struggle to systematically address these complex, end-to-end intelligent generation tasks.

Inspired by agentic technologies, researchers have begun to construct image agentic systems. Commercial products such as Seedream~\cite{seedream2025seedream}, Nano Banana Pro~\cite{deepmind2025nanobananapro}, and FLUX-2 Max~\cite{flux2max} have already demonstrated the immense potential of synergizing search and reasoning in visual generation tasks. Concurrently, the open-source community is actively advancing in this direction; research works like Mind-Brush~\cite{he2026mindbrush} and Unify-Agent~\cite{chen2026unify-agent} leverage the deep reasoning capabilities of Multimodal Large Language Models (MLLMs) alongside external search tools to augment foundational models. Nevertheless, existing open-source solutions either fall short in their capacity to handle complex problems or require prohibitively expensive data and training costs, failing to universally empower a broad spectrum of open-source image models. Consequently, a substantial gap persists between the open-source community and commercial products within the realm of "intelligent generation."

Inspired by the recent paradigm of OpenClaw~\cite{openclaw2026}, we propose RS-Gen, a multi-stage image agentic framework augmented by reasoning and search. RS-Gen diverges from treating image generation and editing as a simplistic black-box mapping process; instead, it reconstructs it into a "questioning-and-solving" closed-loop task driven collaboratively by multi-stage and multi-agent systems. As a plug-and-play, training-free universal solution, RS-Gen can be seamlessly integrated with existing open-source image models, significantly enhancing their capabilities in implicit intent parsing, complex logical reasoning, and real-time information perception. As illustrated in Figure~\ref{fig:show_case}, RS-Gen demonstrates exceptional performance in tasks that necessitate real-time knowledge retrieval and complex logical reasoning. For instance, when confronted with highly challenging prompts such as "rendering the latest concept sports car of a certain brand," "inferring and generating the shape represented by the question mark," and "moving a single matchstick to correct the equation," the framework consistently produces accurate and logically coherent visual content.

Specifically, the core advantages of RS-Gen include:

\begin{enumerate}
    \item \textbf{Implicit Intent Parsing in Multi-Turn Interactions:} Leveraging the agent's memory mechanism and the reasoning capabilities of multimodal models, the system accurately captures the user's deep intentions across multi-turn dialogues. By resolving coreferences and explicitly identifying the source images and target objects for editing, it reconstructs ambiguous and implicit user intentions into clear, actionable image generation and editing instructions.
    \item \textbf{Knowledge Gap and Logical Issue Detection:} Utilizing Multimodal Large Language Models (MLLMs), the system pre-evaluates the complexity of user instructions to precisely identify latent information deficits and logical reasoning barriers. For complex tasks, the system proactively formulates questions centered around these knowledge gaps and logical issues, thereby guiding the subsequent search and reasoning processes.
    \item \textbf{Retrieval-Augmented Generation:} By integrating powerful external information retrieval tools, RS-Gen thoroughly breaks the static knowledge constraints imposed by the models' inherent training data, providing accurate factual grounding and a robust logical basis for the subsequent image generation and editing.
    \item \textbf{Autonomous Planning and Self-Correction Mechanism Driven by the ReAct Paradigm:} Drawing upon the ReAct~\cite{yao2022react} pattern, the system follows an "Observation-Thought-Action" paradigm prior to executing specific image generation. It autonomously conducts step-by-step planning and tool invocation, dynamically adjusting strategies based on tool feedback to ensure that knowledge gaps and logical reasoning issues are resolved before image generation begins. Concurrently, during the image generation phase, the system introduces a "Generate-Review-Correct" self-correction closed loop, which significantly enhances the reliability and robustness of the final output.
\end{enumerate}

\section{Related Work}

\subsection{Image Generation and Editing}
In recent years, Latent Diffusion Models (LDMs), prominently represented by Stable Diffusion (SD)~\cite{rombach2021stable-diffusion}, have achieved milestone advancements in the field of image generation. By performing iterative denoising within a low-dimensional latent space, SD enables the efficient synthesis of high-resolution images. Subsequently, pioneering works such as SD3~\cite{esser2024sd3} and FLUX.1~\cite{flux2024} introduced Flow Matching~\cite{lipman2022flowmatch} techniques. Coupled with large-scale Transformer~\cite{vaswani2017attention} architectures, these models have further elevated the fidelity and photorealism of visual generation to unprecedented heights. However, despite these massive leaps in visual synthesis quality, such models continue to confront a significant "semantic gap" when tackling complex tasks. Their underlying architectures rely heavily on static pre-trained text encoders, such as CLIP~\cite{radford2021clip} or T5~\cite{raffel2020t5-text-encoder}. Constrained by limited semantic representation spaces, these models frequently exhibit a severe degradation in instruction understanding and instruction following capabilities when processing complex prompts that involve ambiguous intentions or coreferences. Inherently, the generation process of these models remains confined to a "text-to-pixel" statistical mapping, lacking the capacity for deep comprehension and logical reasoning regarding the high-level semantics underlying complex instructions.

As user demands evolve from basic "generation from scratch" to the fine-grained modification of existing visual content, image editing has emerged as a highly challenging frontier research topic. Unlike pure image generation tasks, image editing is confronted with far more stringent multi-dimensional constraints: the model must not only accurately parse and execute editing instructions but also strictly preserve the semantic coherence and visual textural consistency of non-editing regions during the editing process. In this context, InstructPix2Pix~\cite{brooks2022instructpix2pix} pioneered a data-driven "instruction-to-edit" mapping paradigm. Subsequently, a series of representative works, such as MagicBrush~\cite{Zhang2023MagicBrush}, HQ-Edit~\cite{hui2024hqedit}, and Emu Edit~\cite{sheynin2024emuedit}, have successively emerged. These approaches primarily fine-tune text-to-image models by constructing large-scale triplet datasets—comprising the source image, editing instruction, and target image—which has significantly enhanced the models' editing precision and generalization capabilities across fine-grained tasks, including local modifications, style transfers, and attribute adjustments.

To further mitigate the challenges of complex instruction comprehension, researchers have begun exploring the integration of Vision-Language Models (VLMs) into image editing architectures. This approach aims to leverage the powerful multimodal understanding and logical reasoning capabilities of VLMs, thereby enhancing the semantic parsing, context awareness, and fine-grained control capabilities of image editing models. For instance, works such as OmniGen2~\cite{wu2025omnigen2}, Qwen-Image-Edit~\cite{wu2025qwenimagetechnicalreport}, and Step1X-Edit~\cite{liu2025step1x-edit} innovatively employ VLMs to replace traditional text encoders like CLIP~\cite{radford2021clip} or T5~\cite{raffel2020t5-text-encoder}, utilizing the deep semantic representations output by VLMs as control conditions for the generation process. The evolution of such architectural paradigms has significantly elevated the models' comprehension capabilities when tackling long texts and complex instructions.

However, although the introduction of VLMs has endowed image generation and editing models with enhanced semantic comprehension capabilities, such architectures still confront the following three fundamental bottlenecks when tackling real-world and complex user tasks:

\begin{enumerate}
    \item \textbf{Coreference Ambiguity in Multi-Turn Interactions:} Real-world application scenarios typically involve continuous multi-turn interactions, where user instructions are often highly colloquial and contain implicit referential pronouns (e.g., "add a cat next to him," "change it to a dog instead"). Existing image models lack contextual understanding and memory mechanisms, making it difficult to accurately parse user intentions and pinpoint the target objects for editing across multiple dialogue turns. Such intention resolution failures easily lead to unsuccessful image editing, rendering them inadequate to support real and complex user demands.
    \item \textbf{Static Knowledge Cutoff and Hallucination:} For the vast majority of models, their parameterized knowledge boundaries are solidified upon the completion of training. When confronted with out-of-distribution (OOD) novel concepts, real-time information, or long-tail entities (e.g., "generate the newly released conceptual supercar by brand X"), models frequently suffer from severe visual and factual hallucinations due to the lack of prior knowledge regarding the target's accurate appearance and detailed features. This results in generated content that is severely decoupled from the objective real world.
    \item \textbf{Lack of Explicit Logical Reasoning Mechanisms:} Inherently, the generation process of existing image models remains a "text-to-pixel" statistical mapping, typically requiring users to provide direct and explicit instructions. When faced with instructions that possess ambiguous intentions or implicitly embed complex physical laws or visual logic puzzles (e.g., "draw the scene two hours later," "deduce and draw the shape represented by the question mark based on the visual patterns in the image"), models are incapable of performing explicit deep reasoning. Lacking the ability for in-depth deconstruction of the underlying rules behind the instructions and visual content, the generated results often violate physical common sense and basic logic, thereby failing to fulfill the users' advanced demands.
\end{enumerate}

\subsection{Unified Multimodal Understanding and Generation Models}
To fundamentally break the barriers between understanding and generation tasks and achieve their deep synergy, researchers have proposed Unified Multimodal Understanding and Generation Models. Unlike the traditional separated paradigm of "text encoder + image generator," unified architectures aim to map visual understanding and visual generation into the same representation space for joint modeling. Works such as Chameleon~\cite{Chameleon_Team_Chameleon_Mixed-Modal_Early-Fusion_2024}, Show-o~\cite{xie2024show-o}, Janus~\cite{wu2024janus}, and Bagel~\cite{deng2025bagel} have fully demonstrated the immense potential of the native unification of understanding and generation in enhancing generative performance. By leveraging their native and powerful multimodal perception and parsing capabilities, such models can more profoundly comprehend physical common sense and logical rules, thereby significantly alleviating the issue of "visual hallucinations" that violate objective common sense in the generated results.

However, although unified models have achieved significant progress in multimodal understanding and synergistic generation, they still confront severe challenges when tackling high-complexity tasks: (1) Static Knowledge and Factual Hallucinations: The capability boundaries of unified models are constrained by the static data distribution during the training phase. Once model training is completed, it inherently faces the "knowledge cutoff" issue. When confronted with continuously emerging novel concepts and new knowledge in the real world, such models inevitably produce factual hallucinations, making it difficult to guarantee the accuracy of the generated content. (2) Prohibitive Training Costs: Unified models rely on massive amounts of high-quality interleaved image-text data and large-scale computing power. This renders their training costs prohibitively high, making it difficult for the open-source community to bear their adaptation and iteration costs. (3) Lack of Explicit Multi-Step Planning and Deep Reasoning Mechanisms: Despite outperforming traditional separated architectures at the multimodal perception level, unified models are unable to provide highly reliable and logically rigorous generated results when dealing with advanced tasks requiring multi-step planning, self-correction, and deep reasoning, due to their inherently limited implicit reasoning capabilities.

\subsection{Visual Agentic Systems}
To break through the performance bottlenecks of monolithic architectures in complex multimodal understanding and generation tasks, the field of artificial intelligence has recently been undergoing a paradigm shift from traditional "instruction-following" models to "AI Agents" equipped with autonomous decision-making capabilities. Modern agents are no longer confined to unimodal pure-text interactions or basic question-answering; rather, they have evolved into composite architectures integrating advanced capabilities such as multimodal perception, long- and short-term memory, autonomous multi-step planning, and external tool invocation.

Within the agentic architecture, the ReAct~\cite{yao2022react} paradigm endows models with the capability to acquire environmental feedback and adjust strategies in real-time through an explicit "Thought-Action-Observation" closed-loop mechanism, thereby significantly enhancing the logical rigor of agent systems when handling complex tasks. Cutting-edge agent applications, exemplified by OpenClaw~\cite{openclaw2026}, further demonstrate the powerful potential of high-level synergy among foundation models, memory modules, external tools, and specialized skills, successfully transforming ambiguous user demands into executable automated workflows.

Inspired by these advancements, researchers have begun exploring Image Agents specifically tailored for visual tasks, attempting to reconstruct traditional image generation and editing tasks into agent-driven automated workflows. Works such as Mind-Brush~\cite{he2026mindbrush}, Unify-Agent~\cite{chen2026unify-agent}, and Gen-Seacher~\cite{feng2026gen-searcher} attempt to deploy Multimodal Large Language Models (MLLMs) as the core decision-making hub. By invoking external retrieval tools to bridge inherent knowledge gaps, these approaches effectively suppress the generation of factual hallucinations. These efforts mark the evolution of generative AI from pure "text-to-pixel" shallow mapping to agentic architectures equipped with autonomous planning and closed-loop processing capabilities for complex visual tasks.

However, existing image agent frameworks remain in an exploratory stage and confront the following significant challenges when addressing exceptionally complex tasks:

\begin{enumerate}
    \item \textbf{System Coupling and Training Cost:} Existing image agent architectures are frequently deeply bound to specific underlying models or require prohibitively expensive training. For instance, Unify-Agent reframes image generation as an agentic pipeline but requires curating 143K highly tailored multimodal trajectories to explicitly fine-tune the model. This high-coupling and computationally expensive training paradigm severely restricts its generalization capabilities within the open-source ecosystem. Lacking plug-and-play flexibility, it struggles to broadly empower diverse foundation models.
    \item \textbf{Workflow Rigidity and Absence of Fault Tolerance:} Current frameworks typically adopt a linear execution pipeline. For example, Mind-Brush executes a one-way "think-research-create" paradigm. If intermediate retrieval steps fail or the generated image contains flaws, the system lacks an effective self-reflection and dynamic error-correction mechanism. This rigid workflow makes the system highly susceptible to error cascading, ultimately causing the generated results to deviate from factual accuracy or violate logical principles.
\end{enumerate}

To address the aforementioned challenges, we propose the \textbf{RS-Gen} framework. Deeply inspired by the ReAct~\cite{yao2022react} paradigm in the agentic domain, we innovatively construct a multi-stage agentic architecture enhanced by reasoning and search. As a highly flexible, training-free solution, RS-Gen aims to comprehensively break through the performance bottlenecks of existing open-source models when dealing with ambiguous intentions, complex logical reasoning, and knowledge cutoffs through explicit logical deduction chains, factual retrieval closed-loops, and dynamic error-correction mechanisms, thereby significantly elevating the reliability and accuracy of the generated results.

\section{RS-Gen}

\begin{figure}[htbp]
    \centering 
    \includegraphics[width=1.0\textwidth]{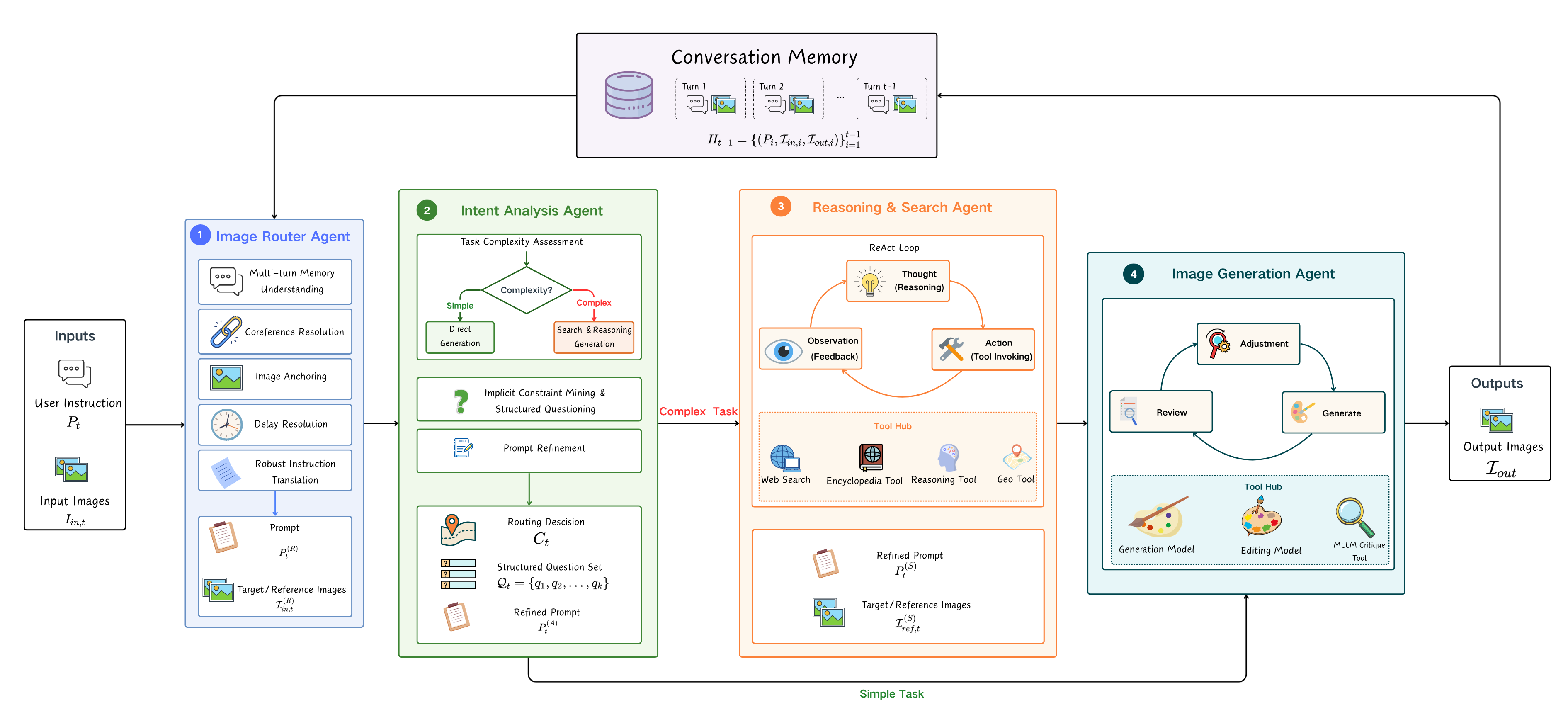} 
    \captionof{figure}{The overall framework of \textbf{RS-Gen}. The overall architecture of the RS-Gen framework. First, the user input is processed by the Image Routing sub-agent to accurately identify target images and disambiguate vague references in the instructions. Then, the Intent Analysis sub-agent evaluates task complexity to plan the optimal execution path, dynamically extracting key sub-problems and refining user instructions as needed. Subsequently, the Reasoning and Search sub-agent autonomously invokes external tools to accomplish factual information retrieval and logical reasoning. Finally, the Image Generation sub-agent executes specific image generation and editing tasks, achieving iterative self-correction through a built-in review mechanism.} 
    \label{fig:rs-gen} 
\end{figure}

We propose \textbf{RS-Gen}, a multi-stage reasoning and search-augmented agentic framework designed to decouple and reconstruct image generation and editing tasks from a monolithic "end-to-end" black-box mapping into a multi-stage collaborative agentic workflow. As illustrated in Figure~\ref{fig:rs-gen}, RS-Gen operates through the synergy of four sub-agents with specifically defined responsibilities. The core design philosophy of this framework is to introduce explicit reasoning chains and closed-loop search feedback mechanisms. Prior to executing specific image generation actions, it pre-emptively resolves ambiguous semantics in user instructions, bridges knowledge gaps, and completes complex logical reasoning at both the logical and knowledge levels. Consequently, it transforms the original inputs into highly precise and informationally complete image generation and editing instructions, as well as reference images.

The execution workflow of \textbf{RS-Gen} strictly adheres to the progressive paradigm of "Perception - Analysis - Retrieval \& Reasoning - Generation". The functional definitions of each core module are as follows:

\begin{enumerate}
    \item \textbf{Image Router Agent:} Serving as the entry module of the system, this agent perceives contextual information through a memory mechanism to accurately comprehend the user's ambiguous task intentions. Its core responsibilities include localizing the target image and specific regions to be edited, and disambiguating vague references in colloquial user instructions, thereby preliminarily transforming original unstructured inputs into structured initial image operation instructions with clear intentions.
    \item \textbf{Intent Analysis Agent:} This agent is responsible for conducting in-depth semantic parsing of the initial instructions. It mines potential requirements for logical reasoning and knowledge gaps, abstracting and distilling them into specific structured queries, which provide clear goal guidance for the downstream retrieval and reasoning stage.
    \item \textbf{Reasoning \& Search Agent:} Acting as the "cognitive hub" of the system, this module draws inspiration from the ReAct~\cite{yao2022react} pattern, possessing the capability to autonomously invoke external search engines, Visual Question Answering (VQA) tools, and logical reasoning tools. Through multi-step deep reasoning and external information retrieval, this agent effectively bridges knowledge blind spots, supplements factual evidence, and accomplishes complex logical deconstruction. Ultimately, it integrates multi-source information to further reconstruct the user's intent into precise instructions and reference images that are informationally complete, logically self-consistent, factually grounded, and compliant with physical laws and objective common sense.
    \item \textbf{Image Generation Agent:} Functioning as the "execution and feedback terminal" of the system, this agent adopts the ReAct~\cite{yao2022react} execution mode and incorporates core tools such as "Generation", "Editing", and "Critique". It proactively executes image generation and editing tasks, and rigorously evaluates the alignment between the generated results and the user's intentions, as well as the overall visual quality. Consequently, it autonomously decides whether to trigger a new round of local modification, iterative polishing, or complete regeneration. Through this "generation-critique-modification" closed-loop mechanism, the quality and reliability of the final generated results are significantly enhanced.
\end{enumerate}

\subsection{Image Router Agent}
Serving as the entry module of the RS-Gen framework, the Image Router Agent aims to address the challenges of intent ambiguity and coreference resolution in multi-turn multimodal dialogue scenarios. Formally, given the user's original instruction $P_t$ and the input image set $\mathcal{I}_{in,t}$ at the current time step $t$, alongside the multimodal historical context $H_{t-1}=\{(P_i,\mathcal{I}_{in,i},\mathcal{I}_{out,i})\}_{i=1}^{t-1}$ (where $P_i$, $\mathcal{I}_{in,i}$, and $\mathcal{I}_{out,i}$ denote the user input instruction, the input image set, and the system output image set at time step $i$, respectively), this agent outputs structured task information. Specifically, this comprises the image operation instruction $P^{(R)}_t$, which has undergone coreference resolution and preliminary optimization, as well as the designated target or reference image set $\mathcal{I}^{(R)}_{in,t}$.
\begin{equation}
    (P^{(R)}_{t}, \mathcal{I}^{(R)}_{in,t}) = \text{F}^{(R)} (P_{t},\mathcal{I}_{in,t},H_{t-1})
    \label{eq:image_route_equation1}
\end{equation}

\textbf{Multimodal Coreference Resolution and Image Anchoring:} In complex multi-turn generation or editing tasks, user instructions frequently contain implicit intents or ambiguous colloquial references (e.g., "draw the result represented by the question mark" or "change its color to red"). Leveraging a memory mechanism, the agent deeply parses the multimodal historical context $H_{t-1}$ to execute precise coreference resolution. For the anchoring of target or reference images, this module designs a hybrid strategy based on time decay and intent matching. Specifically: if the intent is new image generation without specifying or uploading a reference image, the target image set is empty, i.e., $\mathcal{I}^{(R)}_{in,t}=\emptyset$; if the intent is new image generation and a reference image is specified or uploaded, the system accurately identifies and extracts this image set as $\mathcal{I}^{(R)}_{in,t}$; if the intent is image editing, the system prioritizes the results of coreference resolution to anchor the target image; otherwise, it defaults to a backtracking mechanism, extracting the most temporally adjacent historical or uploaded image as $\mathcal{I}^{(R)}_{in,t}$.

\textbf{Principle of Delayed Resolution:} To prevent the model from hallucinating in the absence of external factual support and deep reasoning bases, we propose a "Delayed Resolution" mechanism to strictly demarcate the functional boundaries of this module. During the initial instruction reconstruction phase, the mapping function $F^{(R)}$ is solely responsible for image routing and coreference resolution. When encountering the following scenarios, the model is strictly prohibited from utilizing its internal prior knowledge for ungrounded inference or heuristic completion: (1) External Knowledge: Involving explicit objective entities, proper nouns, specific spatiotemporal coordinates, as well as ambiguous entities, events, knowledge, or concepts. (2) State Evolution: Involving spatiotemporal progression or object state changes induced by external forces. (3) Logical Deduction: Involving mathematical calculations, physical laws, domain-specific knowledge, or complex causal logical inferences. Instead, the system treats these high-level semantic constraints as unresolved variables, preserving them intact within $P^{(R)}_t$. This design effectively defers and transfers all complex decisions requiring deep reasoning and external retrieval to the downstream "Reasoning \& Search Agent," thereby ensuring that the final generated results possess solid factual grounding and logical self-consistency.

\textbf{Robust Instruction Translation Protocol:} Considering the generally weak instruction-following capability of downstream image generation and editing models (e.g., diffusion models) regarding negative semantics, this module incorporates a semantic equivalent translation mechanism when generating the final instruction $P^{(R)}_t$. Specifically, this mechanism translates negative constraints within user instructions into equivalent positive visual descriptions. For instance, it reconstructs a negative state constraint such as "do not keep the stickers" into a positive attribute description like "a solid and clean surface." This transformation significantly mitigates the feature contamination in downstream diffusion models caused by the failure of negative semantic guidance during the generation process.

\subsection{Intent Analysis Agent}
Serving as the task routing and complexity evaluator of the RS-Gen framework, the Intent Analysis Agent receives the instruction $P^{(R)}_t$ and the image set $\mathcal{I}^{(R)}_{in, t}$ from the Image Router module. Its primary function is to evaluate the complexity of the current task to determine whether to adopt the conventional direct generation path or to trigger the downstream Reasoning and Search Agent. When it is determined that the task requires external knowledge or logical support, this agent identifies potential knowledge blind spots and logical issues, accurately extracting the key unresolved questions. Formally, the mapping function $F^{(A)}$ of this agent outputs a structured tuple $(C_t, \mathcal{Q}_t, P^{(A)}_t)$, where $C_t$ denotes the routing decision flag, $\mathcal{Q}_t$ represents the extracted structured query set, and $P^{(A)}_t$ is the reconstructed instruction.
\begin{equation}
    (C_t, \mathcal{Q}_t, P^{(A)}_t) = F^{(A)} (P^{(R)}_t, \mathcal{I}^{(R)}_{in, t}) 
    \label{eq:intent_analysis_equation2}
\end{equation}

\textbf{Task Complexity Assessment and Adaptive Routing:} To ensure an optimal balance between execution efficiency and generation accuracy, this agent introduces a task complexity assessment mechanism aimed at identifying potential challenges of the task in knowledge dimensions (e.g., time-sensitive information, domain-specific knowledge) and logical dimensions (e.g., physical state evolution, visual puzzle-solving). Based on the evaluation results, the system executes the following adaptive routing strategies:
\begin{itemize}
    \item \textbf{Direct Generation ($C_t = \text{direct\_generation}$):} When $P^{(R)}_t$ possesses a clear intent and explicit entities, without knowledge gaps, complex physical law deductions, or deep logical reasoning requirements, the system determines that the internal prior knowledge of the downstream image generation model is sufficient to cover the current task requirements, thereby routing the task directly to the image generation module.
    \item \textbf{Reasoning and Search ($C_t = \text{search\_reason\_generation}$):} When $P^{(R)}_t$ involves time-sensitive constraints, abstract concepts, specific entities lacking visual references, implicit scientific common sense, physical state evolution, or features requiring logical puzzle-solving, the system triggers a deep parsing pipeline and dynamically routes the task to the downstream "Reasoning and Search Agent".
\end{itemize}

\textbf{Mining of Implicit Constraints and Structured Questioning}: When the routing strategy is determined as $C_t = \text{search\_reason\_generation}$, the Intent Analysis Agent deeply parses ambiguous instructions, accurately mining the implicit constraints hidden behind natural language and translating them into an explicit, structured question set $\mathcal{Q}_t = \{q_1, q_2, \dots, q_k\}$ directed at downstream modules. To ensure the completeness of question coverage, this module designs multi-dimensional constraint mining and questioning strategies:

\begin{itemize}
    \item \textbf{Defensive Questioning:} For specific entities, proprietary concepts, or named events involved in the instructions, this module establishes a "fact-evidence first" defensive questioning mechanism. Except for unambiguous basic universal entities (e.g., common flora and fauna, daily objects), even if the downstream image diffusion model contains prior knowledge of the concept, this strategy strictly prohibits the model from relying on its internal prior knowledge for unconstrained heuristic generation. The system forces these non-universal or long-tail entities to be translated into explicit questions, specifically interrogating their exact objective visual features. This mechanism essentially acts as a "cognitive firewall" within the system, fundamentally mitigating the knowledge hallucination issues common in image generation models.
    \item \textbf{Process Decoupling and Terminal Visual Anchoring:} When faced with instructions implying complex dynamic reasoning (e.g., physical laws, biochemical reactions, visual puzzles), this strategy mandates the agent to discard descriptions of intermediate procedural details. By introducing a "process decoupling" mechanism, the structured questions generated by the agent must bypass the convoluted intermediate deduction phases and directly target the final visual features. Specifically, this strategy strictly constrains the agent to focus solely on and specifically interrogate the terminal visual features—such as the final physical morphology and spatial layout of the target entities—that are directly renderable by the diffusion model.
\end{itemize}

\textbf{Modality Fusion and Prompt Refinement:} When a task is routed as direct generation ($C_t = \text{direct\_generation}$), the Intent Analysis Agent bypasses the structured questioning phase (i.e., $\mathcal{Q}_t = \emptyset$) and directly refines the initial instruction $P^{(R)}_t$ to generate an operational prompt $P^{(A)}_t$ that is easily interpretable and renderable by the underlying models. Considering the fundamental mechanistic differences in instruction-following between image generation and editing models, the agent executes an adaptive prompt optimization strategy based on the prior conditions of the input modalities (i.e., the presence or absence of reference images) and the specific task type, ensuring optimal semantic alignment between the final output prompt $P^{(A)}_t$ and the underlying diffusion models. Specifically, this strategy comprises the following three execution paths:
\begin{itemize}
    \item \textbf{High-density Semantic Expansion (for text-only generation tasks):} For pure text-driven generation tasks without reference images, the agent performs high-density semantic expansion, automatically supplementing fine-grained visual descriptions such as environmental background, perspective composition, lighting, materials, and artistic styles to compensate for the information sparsity of the initial text.
    \item \textbf{Terminal Visual Anchoring (for generation tasks with reference images):} For generation tasks accompanied by reference images, the agent strictly adheres to a "terminal-visual-state-oriented" principle. It actively filters out all procedural action descriptions (e.g., operational verbs or intermediate deduction logic) and outputs only absolute objective visual descriptions of the "final desired scene," thereby minimizing the risk of feature confusion in the underlying diffusion models.
    \item \textbf{Structured Action Decomposition (for image editing tasks):} For image editing tasks, the agent translates ambiguous natural language into a strict, structured instruction formatted as "Editing Action + Target Object + Resulting State," guiding the underlying image editing model to execute precise spatial and semantic modifications.
\end{itemize}

\subsection{Reasoning \& Search Agent}
The Reasoning and Search Agent assumes the core functions of complex logical reasoning and external knowledge acquisition within the RS-Gen framework. This module is activated when the intent analysis agent outputs the routing decision $C_{t} = \text{search\_reason\_generation}$. Given the instruction $P_{t}^{(R)}$, the input image set $\mathcal{I}^{(R)}_{in,t}$, the structured question set $\mathcal{Q}_t$, and the tool library $\mathcal{T}$, the agent adopts the ReAct (Reason-Act-Observe) paradigm to execute multiple rounds of external information retrieval and logical reasoning. Ultimately, the agent outputs an information-complete, logically self-consistent, and intent-precise image manipulation instruction $P_{t}^{(S)}$, alongside a high-quality visual reference set $\mathcal{I}^{(S)}_{ref,t}$:
\begin{equation}
    \left( P_{t}^{(S)},\mathcal{I}^{(S)}_{ref,t} \right) = F^{(S)} \left( P_{t}^{(R)},\mathcal{I}^{(R)}_{in,t},\mathcal{Q}_t, \mathcal{T} \right)
    \label{eq:reason_search_equation3}
\end{equation}

\textbf{Reasoning and Search Loop:} For each sub-question $q \in \mathcal{Q}_t$ within the structured question set $\mathcal{Q}_t$, this module draws inspiration from the ReAct paradigm in the agent domain to construct a rigorous "Thought-Action-Observation" reasoning and search loop. The system deploys a diversified expert tool library $\mathcal{T} = \{\tau_{geo}, \tau_{reason}, \tau_{vqa}, \tau_{web}, \tau_{img}\}$, encompassing geographic information querying, deep logical reasoning, visual understanding and recognition, web search engines, and image retrieval engines. Under the ReAct paradigm, the agent breaks through traditional static preset pipelines. Relying on the powerful cognitive capabilities of multimodal large models, it can adaptively invoke external tools to verify hypotheses, supplement critical information, or conduct in-depth reasoning, thereby gradually eliminating information uncertainty. This loop will continue to operate until all implicit logical and knowledge gaps are fully bridged.

\textbf{Cross-modal Cascade Retrieval and Adaptive Fallback:} To address the inherent uncertainties during external tool invocation (e.g., API failures, noisy retrieval results, or information absence), this module designs a robust cross-modal cascade retrieval and adaptive fallback mechanism to ensure system stability:
\begin{itemize}
    \item \textbf{Text-then-Image Cascade Anchoring:} When a target entity involves ambiguous references or broad semantic constraints (e.g., "a certain brand's latest concept car"), the system strictly prohibits directly invoking the image search tool $\tau_{img}$. The agent prioritizes invoking the web search tool $\tau_{web}$ to acquire relevant textual information. After precisely locking onto the entity's objective identifier (e.g., a specific model or professional name), it then utilizes this exact expression as the query to execute the image retrieval. This cascade strategy of "text preceding image" effectively eliminates semantic ambiguity and fundamentally ensures the accuracy of the visual reference images.
    \item \textbf{Multi-round Reformulation and Visual Substitution:} When target image retrieval fails or the returned image quality is substandard (e.g., blurry images, excessive watermarks, mismatch between text and image), the agent automatically triggers the fallback mechanism and sequentially executes the following actions based on priority: switching to alternative search tools, relaxing the semantic constraint boundaries of the search, or seeking surrogate entities with highly similar visual features. This mechanism maximizes the guarantee that, even in extreme scenarios of information absence, the final image generation remains grounded in reliable factual evidence and visual references.
\end{itemize}

\subsection{Image Generation Agent}
The Image Generation Agent is responsible for executing the image operational instructions passed from upstream modules to synthesize the final high-quality images. Unlike the traditional single-step generation paradigm, this module discards the static single-step generation pipeline. Instead, it encapsulates the underlying foundational models for image generation and editing into independent expert tools, upon which it constructs an iterative self-verifying loop based on the ReAct paradigm. Given the generation instruction $P^{(G)}_{t}$ and the reference image set $\mathcal{I}^{(G)}_{ref}$, the agent dynamically invokes the tool library to ultimately output a target image set $\mathcal{I}_{out}$ that is highly aligned with the user's intent. This process can be formally defined as:
\begin{equation}
    \mathcal{I}_{out} = F^{(G)}(P^{(G)}_{t}, \mathcal{I}^{(G)}_{ref}, \mathcal{T}_{gen}) 
    \label{eq:image_generation_equation4}
\end{equation}
where $\mathcal{T}_{gen}$ denotes the expert toolset, encompassing tools for image generation, image editing, and image verification; the instruction $P^{(G)}_{t}$ dynamically maps to either $P^{(A)}_{t}$ from direct generation or $P^{(S)}_{t}$ from the reasoning and search output, depending on the routing decision of the intent analysis; similarly, the reference image set $\mathcal{I}^{(G)}_{ref}$ corresponds to either the initial input $\mathcal{I}^{(R)}_{in,t}$ or the high-quality retrieved reference set $\mathcal{I}^{(S)}_{ref,t}$.

\textbf{Dynamic Tool Invocation and Self-correcting Loop:} The core innovation of this module lies in fundamentally breaking the traditional open-loop generation paradigm by transforming image generation into a multi-round "Generate-Verify-Correct" iterative process. In the $k$-th iteration, the agent executes the following workflow:
\begin{itemize}
    \item \textbf{Image Generation and Editing:} The agent parses the generation strategy $p_k$ for the current round (with the initial state $p_0=P^{(G)}_t$), autonomously and dynamically schedules the required generation or editing models from the expert tool library $\mathcal{T}_{gen}$, and synthesizes the candidate image $I_k$ for the current round.
    \item \textbf{Multimodal Alignment Verification:} Once generated, the candidate image is not output directly. Instead, a visual verification tool based on a Multimodal Large Language Model (MLLM) is invoked to conduct a rigorous review of the image quality and visual-semantic consistency of $I_k$. This step evaluates whether visual elements—such as entity attributes, spatial topological relations, and stylistic features in $I_k$—strictly align with the instruction $P^{(G)}_t$, while simultaneously checking for artifacts or structural flaws, ultimately generating a structured verification report $v_k$.
    \item \textbf{Strategy Adjustment and Image Correction:} If the verification report $v_k$ indicates semantic deviations or visual defects in the current candidate image (e.g., violations of common sense or objective laws), the agent triggers an internal reasoning mechanism. Based on the diagnostic feedback from $v_k$, the agent autonomously deduces targeted correction strategies $\Delta p_k$ and updates the execution parameters for the next round as $p_{k+1} = \text{Update}(p_k, \Delta p_k, P^{(G)}_t)$.
\end{itemize}
The aforementioned "Generate-Verify-Correct" loop continuously and iteratively optimizes the candidate image until the visual verification tool determines that the image achieves complete pixel-level and semantic-level alignment with the user's intent. At this point, the loop terminates and outputs the final result $I_{out}$. This self-correcting mechanism effectively compensates for the instruction-following and visual quality issues prone to occur in traditional open-loop image generation models when processing complex instructions, significantly enhancing the reliability and robustness of the system's output.

\section{Experiments}
\subsection{Benchmarks and Evaluation Protocols}
To comprehensively evaluate the overall capabilities of the RS-Gen framework in complex instruction understanding, knowledge retrieval augmentation, and logical reasoning, we conducted extensive experiments on two core visual tasks: image generation and image editing. For these two tasks, we selected two highly challenging, cutting-edge evaluation benchmarks for in-depth validation:

\begin{table}[htbp]
    \centering
    \caption{Performance of different models on the WISE\_Verified~\cite{niu2025wise} benchmark. The table is categorized into four parts: Commercial Models, Generation-Only Models, Unified Models, and our proposed RS-Gen. The best results within each group are highlighted in bold. A dash ("-") denotes that the evaluation results are not available.}
    \label{tab:wise_result_table}
    
    \small 
    \begin{tabular*}{\textwidth}{@{\extracolsep{\fill}} lccccccc @{}}
        \toprule 
        \textbf{Model} & \textbf{Cultural} & \textbf{Time} & \textbf{Space} & \textbf{Biology} & \textbf{Physics} & \textbf{Chemistry} & \textbf{Overall} \\
        \midrule 
        
        \multicolumn{8}{c}{\textbf{\textit{Commercial Models}}} \\ [0.5ex] 

        Seedream-4.5 & 0.9025 & 0.7500 & 0.8167 & 0.7167 & 0.6917 & 0.7250 & 0.8050 \\
        GPT-Image-1 & 0.8900 & 0.6917 & 0.8833 & 0.8000 & 0.7583 & 0.7750 & 0.8250 \\
        Nano Banana Pro & 0.8975 & 0.8167 & 0.9333 & 0.8167 & 0.8667 & 0.8750 & \textbf{0.8760} \\
        
        \midrule 
        
        \multicolumn{8}{c}{\textbf{\textit{Generation Only Models}}} \\ [0.5ex]
        
        SD-1.5 & 0.4450 & 0.3083 & 0.2083 & 0.2083 & 0.2167 & 0.1500 & 0.3090 \\
        SD-XL-0.9 & 0.4925 & 0.3667 & 0.2417 & 0.2667 & 0.3333 & 0.1833 & 0.3640 \\
        FLUX.1-schnell & 0.4650 & 0.3250 & 0.4667 & 0.2083 & 0.3833 & 0.1000 & 0.3640 \\
        SD-3.5-medium & 0.4825 & 0.3750 & 0.3750 & 0.1833 & 0.3917 & 0.2000 & 0.3760 \\
        SD-3-medium & 0.4700 & 0.4083 & 0.4000 & 0.2000 & 0.3750 & 0.2583 & 0.3850 \\
        FLUX.2-klein-4B & 0.4400 & 0.3667 & 0.4667 & 0.3167 & 0.3917 & 0.3333 & 0.4010 \\
        SD-3.5-large & 0.4900 & 0.4083 & 0.4417 & 0.3000 & 0.3750 & 0.2083 & 0.4040 \\
        FLUX.1-dev & 0.5225 & 0.4000 & 0.5333 & 0.1750 & 0.3750 & 0.2417 & 0.4160 \\
        FLUX.2-klein-9B & 0.4900 & 0.3917 & 0.5500 & 0.3833 & 0.4833 & 0.2250 & 0.4400 \\
        Z-Image & 0.5475 & 0.4667 & 0.5083 & 0.3250 & 0.4750 & 0.1750 & 0.4530 \\
        Qwen-Image-2512 & 0.5950 & 0.4750 & 0.6000 & 0.3500 & 0.4917 & 0.2583 & 0.4990 \\
        Qwen-Image & 0.6275 & 0.5250 & 0.5583 & 0.3417 & 0.4833 & 0.2500 & 0.5100 \\
        FLUX.2-dev & 0.6650 & 0.5667 & 0.6583 & 0.3667 & 0.5250 & 0.3750 & \textbf{0.5650} \\
        
        \midrule
        
        \multicolumn{8}{c}{\textbf{\textit{Unified Models}}} \\ [0.5ex]
        
        Janus-Pro-1B & 0.3050 & 0.2333 & 0.2333 & 0.2167 & 0.3083 & 0.2000 & 0.2650 \\
        Janus-1.3B & 0.3175 & 0.2833 & 0.1833 & 0.2250 & 0.3417 & 0.1833 & 0.2730 \\
        Janus-Pro-7B & 0.3700 & 0.3500 & 0.2833 & 0.2833 & 0.4000 & 0.2333 & 0.3340 \\
        Bagel & 0.4125 & 0.3500 & 0.3083 & 0.2000 & 0.4417 & 0.2583 & 0.3520 \\
        Uniworld-V1 & 0.5150 & 0.4917 & 0.5500 & 0.2250 & 0.4000 & 0.1667 & 0.4260 \\
        HunyuanImage-3.0 & 0.5250 & 0.3917 & 0.4833 & 0.3083 & 0.4500 & 0.2417 & 0.4350 \\
        DeepGen 1.0 & 0.6500 & 0.4100 & 0.7200 & 0.3900 & 0.5900 & 0.4500 & 0.5700 \\
        BAGEL (w/ CoT) & 0.7800 & 0.6333 & 0.5667 & 0.3750 & 0.5500 & 0.5083 & \textbf{0.6280} \\

        \midrule 

        RS-Gen  & \textbf{0.9025} & \textbf{0.8750} & \textbf{0.9083} & \textbf{0.6000} & \textbf{0.6917} & \textbf{0.7750} & \textbf{0.8230} \\
        \bottomrule 
    \end{tabular*}
\end{table}

\textbf{WISE~\cite{niu2025wise}:} Traditional text-to-image evaluation benchmarks primarily focus on surface-level evaluations of "text-image alignment," whereas WISE specifically focuses on assessing a model's ability to integrate world knowledge and understand complex semantics. This benchmark contains 1,000 meticulously constructed test prompts spanning 25 sub-domains, including natural sciences, cultural commonsense, and spatio-temporal reasoning. To overcome the limitations of traditional single metrics, WISE introduces a comprehensive scoring metric, WiScore, which fully quantifies model performance through weighted calculations across the following three key dimensions:
\begin{itemize}
    \item Consistency: Evaluates whether the generated image accurately contains all the key elements implied by the prompt.
    \item Realism: Evaluates whether the image conforms to objective physical laws (e.g., light propagation, gravity) and the true physical properties of materials.
    \item Aesthetic Quality: Comprehensively evaluates the compositional design, color harmony, and overall artistic expressiveness of the image.    
\end{itemize}
Specifically, we utilized the WISE\_Verified benchmark (the official updated version of WISE) and employed Qwen3.5-35B-A3B~\cite{qwen3.5} as the judge model. Meanwhile, in order to conduct a comprehensive comparison with existing solutions, we also used the WISE\_legacy evaluation benchmark (the official old version of WISE) and strictly followed the old evaluation protocol, using gpt-4o-2024-05-13 for evaluation. The results are shown in Table \ref{tab:appendix_wise_legacy_result} of the Appendix.

\textbf{RiseBench~\cite{zhao2025risebench}: }In image editing tasks, existing evaluation benchmarks are mostly confined to simple attribute modifications driven by explicit instructions. In contrast, RiseBench aims to more comprehensively evaluate the editing capabilities of models at the level of deep logical reasoning. This benchmark covers four highly challenging testing directions: temporal reasoning, causal reasoning, spatial reasoning, and logical reasoning. In its evaluation system, it quantitatively assesses the model's comprehensive editing capabilities from three core dimensions: Instruction Reasoning, Appearance Consistency, and Visual Plausibility. Specifically, we employed GPT-4.1-2025-04-14 as the judge model for this benchmark.

\begin{table*}[htbp]
    \centering
    \caption{Performance comparison of various models on the RISEBench~\cite{zhao2025risebench} test set. The table is categorized into three parts: Commercial Models, Open-Source Models, and Image Agentic Systems. The best results within each group are highlighted in bold. A dash ("-") denotes that the evaluation results are not available.}
    \label{tab:risebench_result_table}
    
    \small
    \begin{tabular*}{\textwidth}{@{\extracolsep{\fill}} l | ccc | ccccc @{}}
        \toprule
        \multirow{2}{*}{\textbf{Model Name}} & \multicolumn{3}{c|}{\textbf{Evaluation Sub-Dimensions}} & \multicolumn{5}{c}{\textbf{Accuracy}} \\
        \cmidrule(lr){2-4} \cmidrule(lr){5-9}
        & Instr. Reas. & App. Consis. & Vis. Plaus. & Temporal & Causal & Spatial & Logical & Overall \\
        \midrule
        
        Seedream-4.0 & 58.9 & 67.4 & 91.2 & 17.6 & 13.3 & 11.0 & 7.1 & 12.2 \\
        GPT-Image-1-mini & 54.1 & 71.5 & 93.7 & 25.9 & 31.1 & 33.0 & 9.4 & 25.3 \\
        GPT-Image-1 & 62.8 & 80.2 & 94.9 & 36.5 & 34.4 & 37.0 & 10.6 & 30.0 \\
        Nano Banana & 61.2 & 86.0 & 91.3 & 29.4 & 48.9 & 37.0 & 18.8 & 33.9 \\
        Nano Banana Pro & 77.0 & 85.5 & 94.4 & 43.5 & 63.3 & 48.0 & 37.6 & 48.3 \\
        GPT-Image-2 & 73.8 & 89.3 & 94.9 & 45.9 & 66.7 & 50.0 & 34.1 & 49.4 \\
        GPT-Image-1.5 & 69.7 & 92.5 & 94.9 & 57.6 & 62.2 & 62.0 & 21.2 & \textbf{51.4} \\

        \midrule
        FLUX.1-Canny & 20.2 & 13.1 & 77.5 & 0.0 & 0.0 & 0.0 & 0.0 & 0.0 \\
        HiDream-Edit & 30.3 & 12.6 & 74.9 & 0.0 & 0.0 & 0.0 & 0.0 & 0.0 \\
        EMU2 & 22.6 & 38.2 & 78.3 & 1.2 & 1.1 & 0.0 & 0.0 & 0.5 \\
        OmniGen & 22.0 & 32.6 & 55.3 & 1.2 & 1.1 & 0.0 & 1.2 & 0.8 \\
        Step1X-Edit & 25.1 & 41.5 & 73.5 & 0.0 & 2.2 & 2.0 & 3.5 & 1.9 \\
        Ovis-U1 & 33.9 & 52.7 & 72.9 & 1.1 & 3.3 & 4.0 & 2.4 & 2.8 \\
        FLUX.1-Kontext-Dev & 26.0 & 71.6 & 85.2 & 2.3 & 5.5 & 13.0 & 1.2 & 5.8 \\
        BAGEL & 36.5 & 53.5 & 73.0 & 2.4 & 5.6 & 14.0 & 1.2 & 6.1 \\
        Qwen-Image-Edit-2509 & 37.2 & 66.4 & 86.9 & 4.7 & 11.1 & 17.0 & 2.4 & 9.2 \\
        Gemini-2.0-Flash-pre & 49.9 & 68.4 & 84.9 & 11.8 & 14.4 & 11.0 & 2.4 & 10.0 \\
        BAGEL (w/ CoT) & 45.9 & 73.8 & 80.1 & 5.9 & 17.8 & 21.0 & 1.2 & 11.9 \\
        Gemini-2.0-Flash-exp & 48.9 & 68.2 & 82.7 & 9.4 & 16.7 & 23.0 & 4.7 & 13.9 \\
        Qwen-Image-Edit-2511 & 49.9 & 71.0 & 91.5 & 21.2 & 18.9 & 31.0 & 4.7 & \textbf{19.4} \\
        
        \midrule
              
        Mind-Brush & 61.5 & 79.4 & 86.5 & - & - & - & - & 24.7 \\
        RS-Gen & 69.9 & 91.1 & 93.6 & 36.4 & 50.0 & 40.0 & 29.4 & \textbf{39.1}\\
        \bottomrule
    \end{tabular*}
\end{table*}

\subsection{Experiment Settings}
As a plug-and-play general image generation and editing agent framework, RS-Gen can flexibly integrate different foundational visual models to execute image generation and editing tasks. To ensure the fairness and consistency of the experimental evaluation, we maintained uniform system settings across all evaluation benchmarks. Specifically, we adopt Qwen-Image~\cite{wu2025qwenimagetechnicalreport} as the base image generation model and Qwen-Image-Edit-2511~\cite{wu2025qwenimagetechnicalreport} as the base image editing model. Additionally, Doubao-Seed-2.0-Pro-260215~\cite{seed2026seed2-0} is selected as the core driving model for the agent.

Furthermore, we configured the following core tools in the system's expert tool library:
\begin{itemize}
    \item \textbf{Web Search Tool:} We adopted the Baidu Search API and Google Search API to provide information retrieval services, uniformly setting the maximum number of candidate results returned per single query to 5 (Top-5).
    \item \textbf{Geographic Info Tool:} The LocationIQ API was utilized to provide the system with high-precision reverse geocoding capabilities, supporting complex attribute anchoring related to spatial locations.
    \item \textbf{Encyclopedia Database Tool:} The Wikipedia API was invoked to provide accurate and authoritative entry-level encyclopedic knowledge.
    \item \textbf{Deep Reasoning Tool:} We encapsulated GPT-5.4~\cite{openai2026gpt5-4} as an independent expert tool for deep logical reasoning, dedicated to overcoming complex reasoning challenges.
\end{itemize}

\subsection{Main Results}

Table \ref{tab:wise_result_table} presents the quantitative comparison results of various models on the WISE\_Verified~\cite{niu2025wise} benchmark. The experimental results demonstrate that our proposed RS-Gen framework achieves the best performance in the non-commercial group (comprising Generation-Only Models and Unified Models) with an overall score of 0.8230, establishing a significant margin of 0.195 over the runner-up, BAGEL (w/ CoT)~\cite{deng2025bagel} (0.6280). Even when compared to commercial closed-source models, RS-Gen exhibits strong competitiveness, trailing the top-performing commercial model, Nano Banana Pro~\cite{deepmind2025nanobananapro} (0.8760), by a marginal gap of merely 0.053.
Furthermore, across all fine-grained sub-domains of WISE\_Verified (i.e., Cultural, Time, Space, Biology, Physics, and Chemistry), RS-Gen consistently achieves state-of-the-art (SOTA) performance within the non-commercial category. More importantly, by utilizing Qwen-Image~\cite{wu2025qwenimagetechnicalreport} as the foundational image generation model, the RS-Gen framework yields a substantial absolute score gain of 0.313 over the Qwen-Image baseline (0.5100), successfully elevating its performance to the SOTA level among non-commercial models. This substantial improvement thoroughly validates the exceptional effectiveness of our framework in bridging the knowledge gaps of foundational models and enhancing their logical reasoning capabilities.

Table \ref{tab:risebench_result_table} presents the quantitative comparison results of various models on the RISEBench~\cite{zhao2025risebench} benchmark. The experimental results demonstrate that our proposed RS-Gen framework achieves an overall score of 39.1, reaching the state-of-the-art (SOTA) level among non-commercial models. Compared to the best-performing model in the open-source group, Qwen-Image-Edit-2511~\cite{wu2025qwenimagetechnicalreport} (19.4), RS-Gen establishes a substantial lead of 19.7 points. More importantly, when utilizing Qwen-Image-Edit-2511 as the foundational image editing model. Empowered by RS-Gen's robust information retrieval and logical reasoning capabilities, the model yields a significant performance gain of 19.7 points. Specifically, in the Instruction Reasoning evaluation dimension, the model demonstrates a substantial improvement of 20 points. Furthermore, across complex sub-tasks such as temporal reasoning, causal reasoning, spatial reasoning, and logical reasoning, the RS-Gen framework drives a quantum leap in performance, with the maximum single-item improvement reaching up to 31.1 points. Compared to the Mind-Brush framework (built upon Qwen-Image-Edit-2512) which scores 24.7, RS-Gen achieves a significant lead of 14.4 points. This substantial performance margin empirically validates the structural superiority of RS-Gen. While Mind-Brush relies on a static, open-loop retrieval and generation pipeline, RS-Gen innovatively incorporates an adaptive fallback/fault-tolerance mechanism alongside a "Generate-Review-Correct" closed loop. This architectural design fundamentally empowers RS-Gen with autonomous error-correction capabilities, which is critical for excelling in highly demanding complex editing tasks such as those in RISEBench. Even when compared to commercial models, RS-Gen maintains strong competitiveness, with its overall performance trailing only GPT-Image-1.5~\cite{openai2025gptimage1.5}, GPT-Image-2~\cite{openai2026gptimage2}, and Nano Banana Pro~\cite{deepmind2025nanobananapro}. These results thoroughly validate the effectiveness of the RS-Gen framework in endowing foundational image editing models with complex logical reasoning and refined editing capabilities.

\subsection{Ablation Study}
To rigorously evaluate the effectiveness of the key components within the RS-Gen framework, as well as its overall generalizability and robustness, we designed comprehensive ablation studies. Specifically, employing a control-variable methodology, we systematically quantified the independent contribution of each module to the overall performance by progressively enabling the core components of the framework. Furthermore, to validate the generalizability of the RS-Gen framework, in addition to Qwen-Image-Edit-2511~\cite{wu2025qwenimagetechnicalreport}, we incorporated several mainstream image generation and editing models—including LongCat-Image~\cite{meituan2025longcatimage}, Z-Image, LongCat-Image-Edit~\cite{meituan2025longcatimage}, and SeeDream 4.0~\cite{seedream2025seedream}—as underlying foundation models within our evaluation scope.

\begin{table}[htbp]
    \centering
    \caption{Ablation study of the RS-Gen framework on the WISE\_Verified~\cite{niu2025wise} benchmark. We employ Qwen-Image~\cite{wu2025qwenimagetechnicalreport} as the foundational image generation model and progressively integrate the four core sub-agents of the framework. IR, IA, RS, and IG denote the Image Router Agent, Intent Analysis Agent, Reasoning \& Search Agent, and Image Generation Agent, respectively, while QI represents Qwen-Image~\cite{wu2025qwenimagetechnicalreport}. The best results in each column are highlighted in bold.}
    \label{tab:ablation_study_table3}
    
    \small 
    \begin{tabular*}{\textwidth}{@{\extracolsep{\fill}} lccccccc @{}}
        \toprule 
        \textbf{Model} & \textbf{Cultural} & \textbf{Time} & \textbf{Space} & \textbf{Biology} & \textbf{Physics} & \textbf{Chemistry} & \textbf{Overall} \\
        \midrule 
        
        QI & 0.6275 & 0.5250 & 0.5583 & 0.3417 & 0.4833 & 0.2500 & 0.5100 \\
        IR + QI & 0.6750 & 0.6000 & 0.7000 & 0.3333 & 0.4417 & 0.2583 & 0.5500 (+0.04) \\
        IR+IA+QI & 0.6825 & 0.6417 & 0.7333 & 0.4333 & 0.5417 & 0.2500 & 0.5850 (+0.035) \\
        IR+IA+RS+QI & 0.8775 & 0.8667 & 0.8833 & 0.5417 & \textbf{0.7000} & 0.7000 & 0.7940 (+0.209) \\
        IR+IA+RS+IG+QI & \textbf{0.9025} & \textbf{0.8750} & \textbf{0.9083} & \textbf{0.6000} & 0.6917 & \textbf{0.7750} & \textbf{0.8230} (+0.029) \\
        \bottomrule 
    \end{tabular*}
\end{table}

The experimental results in Table \ref{tab:ablation_study_table3} clearly demonstrate that all sub-agents within the RS-Gen framework contribute substantially to the final system performance. Specifically, compared to the Qwen-Image baseline model, merely introducing the Image Router Agent for preliminary prompt optimization improves the overall WISE\_Verified score from 0.5100 to 0.5500. Building upon this, the addition of the Intent Analysis Agent further increases the score to 0.5850. Notably, following the integration of the Reasoning \& Search Agent, the system experiences a significant performance leap, with the overall score substantially increasing to 0.7940. This result compellingly proves that incorporating external knowledge retrieval and deep reasoning capabilities can overcome the cognitive limitations of foundation models, thereby significantly enhancing image generation accuracy. Furthermore, while the first three modules yield considerable performance gains, a single generation pass struggles to entirely avoid flaws or errors. By enabling the Image Generation Agent, the system constructs a self-correcting generation closed-loop, elevating the overall score further to 0.8230. This fully validates the effectiveness of the self-correction mechanism in autonomously identifying and rectifying generation defects.

\begin{table*}[htbp]
    \centering
    \caption{Ablation study of the RS-Gen framework on the RISEBench~\cite{zhao2025risebench} benchmark. We employ Qwen-Image-Edit-2511 as the foundational image editing model and progressively integrate the four core sub-agents of the framework. IR, IA, RS, and IG denote the Image Router Agent, Intent Analysis Agent, Reasoning \& Search Agent, and Image Generation Agent, respectively,  while QIE represents Qwen-Image-Edit-2511~\cite{wu2025qwenimagetechnicalreport}. The best results in each column are highlighted in bold.}
    \label{tab:ablation_study_table4}
    
    \small
    \begin{tabular*}{\textwidth}{@{\extracolsep{\fill}} l | ccc | ccccc @{}}
        \toprule
        \multirow{2}{*}{\textbf{Model Name}} & \multicolumn{3}{c|}{\textbf{Evaluation Sub-Dimensions}} & \multicolumn{5}{c}{\textbf{Accuracy}} \\
        \cmidrule(lr){2-4} \cmidrule(lr){5-9}
        & Instr. Reas. & App. Consis. & Vis. Plaus. & Temporal & Causal & Spatial & Logical & Overall \\
        \midrule

        QIE & 49.9 & 71.0 & 91.5 & 21.2 & 18.9 & 31.0 & 4.7 & 19.4 \\
        IR+QIE & 51.5 & 79.8 & 90.5 & 17.6 & 27.7 & 29.0 & 12.9 & 22.2 (+2.8) \\
        IR+IA+QIE & 56.7 & 81.0 & 92.0 & 21.1 & 38.8 & 35.0 & 10.5 & 26.9 (+4.7) \\
        IR+IA+RS+QIE & 66.4 & 87.1 & 91.1 & \textbf{37.6} & 48.8 & 36.0 & 17.6 & 35.2 (+8.3) \\
        IR+IA+RS+IG+QIE & \textbf{69.9} & \textbf{91.1} & \textbf{93.6} & 36.4 & \textbf{50.0} & \textbf{40.0} & \textbf{29.4} & \textbf{39.1} (+3.9) \\

        \bottomrule
    \end{tabular*}
\end{table*}

The ablation study results on RISEBench~\cite{zhao2025risebench} in Table \ref{tab:ablation_study_table4} further validate the effectiveness of the RS-Gen framework for image editing tasks. Compared to the Qwen-Image-Edit-2511~\cite{wu2025qwenimagetechnicalreport} baseline (overall score of 19.4), progressively integrating each agent module yields consistent performance gains. Specifically, the sequential introduction of the Image Router Agent and the Intent Analysis Agent improves the overall score to 22.2 (+2.8) and 26.9 (+4.7), respectively. Most notably, the integration of the Reasoning \& Search Agent results in the most significant performance leap, bringing the overall score to 35.2 (+8.3). An analysis of the sub-dimensions reveals substantial growth in areas requiring deep comprehension, such as Temporal and Causal reasoning. This compellingly demonstrates that incorporating external knowledge retrieval and complex logical reasoning capabilities can significantly compensate for the foundation model's shortcomings in handling intricate editing instructions. Finally, enabling the Image Generation Agent closed-loop pushes the overall score to 39.1 (+3.9) and achieves optimal performance across multiple core metrics, including Instruction Reasoning (Instr. Reas.) and Visual Plausibility (Vis. Plaus.). These empirical results fully confirm the rationality of the module design within the RS-Gen framework and their synergistic effects.

\begin{table}[htbp]
    \centering
    \caption{Experimental results of the RS-Gen framework integrated with different foundational image generation models on the WISE\_Verified~\cite{niu2025wise} benchmark. This evaluation is designed to demonstrate the cross-model generalizability of the proposed agent framework.}
    \label{tab:ablation_study_table5}
    
    \small 
    \begin{tabular*}{\textwidth}{@{\extracolsep{\fill}} lccccccc @{}}
        \toprule 
        \textbf{Model} & \textbf{Cultural} & \textbf{Time} & \textbf{Space} & \textbf{Biology} & \textbf{Physics} & \textbf{Chemistry} & \textbf{Overall} \\
        \midrule 
        
        LongCat-Image & 0.6850 & 0.4583 & 0.6333 & 0.3667 & 0.4917 & 0.3500 & 0.5500 \\
        LongCat-Image + RS-Gen & 0.8625 & 0.8583 & 0.8417 & 0.6167 & 0.6583 & 0.7167 & 0.7880 (+0.238) \\
        
        \midrule
        
        Z-Image & 0.5475 & 0.4583 & 0.4583 & 0.3583 & 0.4250 & 0.1333 & 0.4390 \\
        Z-Image + RS-Gen & 0.8825 & 0.9000 & 0.9000 & 0.7083 & 0.7500 & 0.8333 & 0.8440 (+0.405) \\

        \bottomrule 
    \end{tabular*}
\end{table}

\begin{table*}[!htbp]
    \centering
    \caption{Experimental results of the RS-Gen framework integrated with different foundational image editing models on the RISEBench~\cite{zhao2025risebench} benchmark. This evaluation is designed to demonstrate the cross-model adaptability and generalizability of the proposed framework in image editing tasks. LCIE denotes LongCat-Image-Edit.}
    \label{tab:ablation_study_table6}
    
    \small
    \begin{tabular*}{\textwidth}{@{\extracolsep{\fill}} l | ccc | ccccc @{}}
        \toprule
        \multirow{2}{*}{\textbf{Model Name}} & \multicolumn{3}{c|}{\textbf{Evaluation Sub-Dimensions}} & \multicolumn{5}{c}{\textbf{Accuracy}} \\
        \cmidrule(lr){2-4} \cmidrule(lr){5-9}
        & Instr. Reas. & App. Consis. & Vis. Plaus. & Temporal & Causal & Spatial & Logical & Overall \\
        \midrule

        LCIE & 37.2 & 74.6 & 85.2 & 4.7 & 12.2 & 12.0 & 4.7 & 8.6 \\
        LCIE + RS-Gen & 63.0 & 86.8 & 87.7 & 37.6 & 40.0 & 25.0 & 14.1 & 29.1 (+20.5) \\
        
        \midrule

        Seedream-4.0 & 58.9 & 67.4 & 91.2 & 17.6 & 13.3 & 11.0 & 7.1 & 12.2 \\
        Seedream-4.0 + RS-Gen & 69.0 & 89.4 & 93.8 & 41.1 & 52.2 & 34.0 & 27.0 & 38.6 (+26.4) \\
    
        \bottomrule
    \end{tabular*}
\end{table*}

To comprehensively evaluate the cross-model generalizability of the RS-Gen framework, we extended our experiments beyond the Qwen series to include various representative open-source and closed-source image generation and editing models as underlying foundation models. As shown in Table \ref{tab:ablation_study_table5}, on the WISE\_Verified~\cite{niu2025wise} benchmark, the RS-Gen framework yields absolute performance gains of +0.238 and +0.405 for LongCat-Image~\cite{meituan2025longcatimage} and Z-Image~\cite{team2025zimage}, significantly boosting their overall scores to 0.7880 and 0.8440, respectively. Furthermore, the RISEBench results in Table \ref{tab:ablation_study_table6} demonstrate exceptional adaptability in complex image editing tasks: upon integrating our framework, the overall scores of LongCat-Image-Edit~\cite{meituan2025longcatimage} and SeeDream-4.0~\cite{seedream2025seedream} surge to 29.1 (+20.5) and 38.6 (+26.4), respectively, achieving a more than threefold performance increase over their baselines. These experimental results robustly confirm that the RS-Gen framework not only effectively overcomes the performance bottlenecks of individual foundation models but also exhibits strong cross-model generalizability and system robustness.

\begin{table}[!htbp]
    \centering
    \caption{Performance comparison of the RS-Gen framework on the WISE\_Verified~\cite{niu2025wise} benchmark across various multimodal models. This experiment aims to evaluate the generalization performance of the RS-Gen framework when employing different multimodal models as core drivers. Mimo and Seed denote MiMo-v2-Omni~\cite{xiaomi2026mimov2omni} and Doubao-Seed-2.0-Pro-260215~\cite{seed2026seed2-0}, respectively.}
    \label{tab:ablation_study_table7}
    
    \small 
    \begin{tabular*}{\textwidth}{@{\extracolsep{\fill}} lccccccc @{}}
        \toprule 
        \textbf{Model} & \textbf{Cultural} & \textbf{Time} & \textbf{Space} & \textbf{Biology} & \textbf{Physics} & \textbf{Chemistry} & \textbf{Overall} \\
        \midrule 
        
        Qwen-Image & 0.6275 & 0.5250 & 0.5583 & 0.3417 & 0.4833 & 0.2500 & 0.5100 \\
        Qwen-Image + RS-Gen(MiMo) & 0.8675 & 0.8333 & 0.8417 & 0.6917 & 0.7250 & 0.7833 & 0.8120(+0.3020) \\
        Qwen-Image + RS-Gen(Seed) & 0.9025 & 0.8750 & 0.9083 & 0.6000 & 0.6917 & 0.7750 & 0.8230(+0.3130) \\
        \bottomrule 
    \end{tabular*}
\end{table}

\begin{table*}[!htbp]
    \centering
    \caption{Performance comparison of the RS-Gen framework on the RISEBench~\cite{zhao2025risebench} benchmark across various multimodal models. This experiment aims to evaluate the generalization performance of the RS-Gen framework when employing different multimodal models as core drivers. QIE, Mimo and Seed denote Qwen-Image-Edit-2511~\cite{wu2025qwenimagetechnicalreport}, MiMo-v2-Omni~\cite{xiaomi2026mimov2omni} and Doubao-Seed-2.0-Pro-260215~\cite{seed2026seed2-0}, respectively.}
    \label{tab:ablation_study_table8}
    
    \small
    \begin{tabular*}{\textwidth}{@{\extracolsep{\fill}} l | ccc | ccccc @{}}
        \toprule
        \multirow{2}{*}{\textbf{Model Name}} & \multicolumn{3}{c|}{\textbf{Evaluation Sub-Dimensions}} & \multicolumn{5}{c}{\textbf{Accuracy}} \\
        \cmidrule(lr){2-4} \cmidrule(lr){5-9}
        & Instr. Reas. & App. Consis. & Vis. Plaus. & Temporal & Causal & Spatial & Logical & Overall \\
        \midrule

        QIE & 49.9 & 71.0 & 91.5 & 21.2 & 18.9 & 31.0 & 4.7 & 19.4 \\
        QIE + RS-Gen (MiMo) & 68.4 & 84.7 & 89.8 & 34.1 & 32.2 & 36.0 & 27.0 & 32.5(+13.1) \\
        QIE + RS-Gen (Seed) & 69.9 & 91.1 & 93.6 & 36.4 & 50.0 & 40.0 & 29.4 & 39.1(+19.7) \\
    
        \bottomrule
    \end{tabular*}
\end{table*}

To comprehensively evaluate the generalization performance of the RS-Gen framework across different core driving models, we further introduce MiMo-v2-Omni~\cite{xiaomi2026mimov2omni} as the core driving model for each sub-agent, alongside Doubao-Seed-2.0-Pro-260215~\cite{seed2026seed2-0}.
As reported in Table ~\ref{tab:ablation_study_table7}, on the WISE\_Verified~\cite{niu2025wise} benchmark, compared to the baseline Qwen-Image~\cite{wu2025qwenimagetechnicalreport} (with an overall score of 0.5100), the RS-Gen framework integrated with MiMo-v2-Omni~\cite{xiaomi2026mimov2omni} and Doubao-Seed-2.0-Pro-260215~\cite{seed2026seed2-0} as core drivers exhibits substantial performance gains, boosting the overall results by 0.3020 and 0.3130 (reaching 0.8120 and 0.8230, respectively). Notably, despite utilizing different foundational models, the ultimate performance profiles are remarkably close, with a marginal gap of only 0.0110. This firmly demonstrates that the RS-Gen framework possesses exceptional robustness in foundational image generation tasks, enabling seamless adaptation to diverse underlying driving models.
On the more challenging RISEBench~\cite{zhao2025risebench} image editing benchmark, the RS-Gen framework consistently delivered stable performance enhancements. As shown in Table ~\ref{tab:ablation_study_table8}, compared to the baseline Qwen-Image-Edit-2511~\cite{wu2025qwenimagetechnicalreport} (19.4), configurations incorporating MiMo-v2-Omni~\cite{xiaomi2026mimov2omni} and Doubao-Seed-2.0-Pro-260215~\cite{seed2026seed2-0} achieve absolute improvements of 13.1 (reaching 32.5) and 19.7 (reaching 39.1), respectively. Crucially, in the Logical reasoning dimension, where the baseline model struggles at a mere 4.7, its integration with RS-Gen drastically elevates the score to 29.4. This underscores that our framework significantly compensates for the inherent deficiencies of vanilla models in complex reasoning. Furthermore, since RISEBench~\cite{zhao2025risebench} imposes higher demands on Logical and Causal reasoning, the superior model (Seed) exhibits stronger explosive potential in this context. This further validates that RS-Gen enjoys excellent scalability, capable of unlocking greater efficacy as the capabilities of the underlying driver models advance.
In summary, consistent empirical results across both benchmarks demonstrate that the RS-Gen framework is model-agnostic, maintaining a high standard of generalization and yielding substantial performance gains across diverse underlying drivers.

\subsection{Limitations \& Future Work}

In this study, we honestly acknowledge that the significant performance gains achieved by the RS-Gen framework largely benefit from the meticulous design of system prompt engineering (e.g., multi-agent role constraints and dynamic feedback mechanism guidance). Although cross-benchmark experiments (such as Table 7 and Table 8) have confirmed that this prompt architecture possesses excellent generalization and robustness across different multimodal baseline models, it also indicates that the current system relies to some extent on the instruction-following capability of the underlying large models. When connected to lightweight or early multimodal models with weaker instruction comprehension capabilities, the expected gains of the framework may be limited.

In response to the above limitations, we plan to conduct in-depth exploration and expansion in future work from the following directions:

\begin{itemize}
    \item \textbf{Adaptive Prompt Evolution:} Future research will introduce an Automatic Prompt Optimizer, enabling the framework to dynamically and autonomously adjust and evolve its prompt strategies during the inference stage according to the characteristics of different underlying multimodal models, thereby further reducing manual design costs and improving the end-to-end automation level and universality of the system.
    \item \textbf{Reflective Skill Library Expansion:} In the existing RS-Gen framework, the agent's reasoning for each task is "independent". In the future, we plan to introduce a training-free experience retrieval mechanism: when the multi-agent successfully tackles a high-difficulty image generation task, the system will automatically distill its prompt strategy and reflection trajectory, consolidating them as a "Skill" into an external database. Faced with similar challenges in the future, the agent can directly reuse successful experience through retrieval, achieving system-level self-evolution and lifelong learning (Lifelong Learning) capabilities.
\end{itemize}

\section{Conclusion}
We propose RS-Gen, a multi-stage image agent framework augmented by reasoning and search. RS-Gen reconstructs image generation and editing into a multi-stage, multi-agent collaborative "problem-posing and problem-solving" closed-loop workflow. As a highly flexible and training-free solution, RS-Gen effectively breaks through the performance bottlenecks of existing open-source models when handling ambiguous intents, complex logical reasoning, and knowledge cutoffs through explicit logical reasoning chains, factual retrieval closed loops, and dynamic self-correction mechanisms, thereby significantly enhancing the reliability and accuracy of the generated results. Experimental results demonstrate that the RS-Gen framework not only endows foundational models with robust logical reasoning capabilities but also effectively bridges their intrinsic knowledge gaps. We believe that the traditional end-to-end single-model generation paradigm will gradually evolve towards agent-based collaborative systems, and Image Agents are poised to become the mainstream direction in the future of image generation and editing.

\section{Acknowledgments}
This paper uses the WISE and RISEBench datasets. The authors confirm that the use of these datasets in this paper is limited to academic research purposes and has not been used for any commercial activities.

\bibliographystyle{unsrtnat} 
\bibliography{references}  






\clearpage
\appendix

\section{Additional Experimental Results}

\begin{table}[htbp]
    \centering
    \caption{Performance comparison of various models on the WISE\_legacy test suite. According to the official leaderboard criteria of WISE\_legacy, this table is structurally categorized into two primary paradigms: dedicated text-to-image models (Dedicated T2I) and unified multimodal large language models (Unified MLLM). Additionally, the benchmark results of representative visual generation agent-based methods are incorporated for a comprehensive evaluation.}
    \label{tab:appendix_wise_legacy_result}
    
    \small 
    \begin{tabular*}{\textwidth}{@{\extracolsep{\fill}} lccccccc @{}}
        \toprule 
        \textbf{Model} & \textbf{Cultural} & \textbf{Time} & \textbf{Space} & \textbf{Biology} & \textbf{Physics} & \textbf{Chemistry} & \textbf{Overall} \\
        \midrule 
        
        \multicolumn{8}{c}{\textbf{\textit{Dedicated T2I}}} \\ [0.5ex] 
        SD-v1-5 & 0.34 & 0.35 & 0.32 & 0.28 & 0.29 & 0.21 & 0.32 \\
        SD-2-1 & 0.30 & 0.38 & 0.35 & 0.33 & 0.34 & 0.21 & 0.32 \\
        FLUX.1-schnell & 0.39 & 0.44 & 0.50 & 0.31 & 0.44 & 0.26 & 0.40 \\
        SD-3-medium & 0.42 & 0.44 & 0.48 & 0.39 & 0.47 & 0.29 & 0.42 \\
        SD-XL-base-0.9 & 0.43 & 0.48 & 0.47 & 0.44 & 0.45 & 0.27 & 0.43 \\
        SD-3.5-Medium & 0.43 & 0.50 & 0.52 & 0.41 & 0.53 & 0.33 & 0.45 \\
        SD-3.5-Large & 0.44 & 0.50 & 0.58 & 0.44 & 0.52 & 0.31 & 0.46 \\
        PixArt-Alpha & 0.45 & 0.50 & 0.48 & 0.49 & 0.56 & 0.34 & 0.47 \\
        playground-v2.5 & 0.49 & 0.58 & 0.55 & 0.43 & 0.48 & 0.33 & 0.49 \\
        FLUX.1-dev & 0.48 & 0.58 & 0.62 & 0.42 & 0.51 & 0.35 & 0.50 \\
        
        \midrule 
        
        \multicolumn{8}{c}{\textbf{\textit{Unify MLLM}}} \\ [0.5ex]

        Orthus-7B-base & 0.07 & 0.10 & 0.12 & 0.15 & 0.15 & 0.10 & 0.10 \\
        JanusFlow-1.3B & 0.13 & 0.26 & 0.28 & 0.20 & 0.19 & 0.11 & 0.18 \\
        Janus-1.3B & 0.16 & 0.26 & 0.35 & 0.28 & 0.30 & 0.14 & 0.23 \\
        Janus-Pro-1B & 0.20 & 0.28 & 0.45 & 0.24 & 0.32 & 0.16 & 0.26 \\
        Orthus-7B-instruct & 0.23 & 0.31 & 0.38 & 0.28 & 0.31 & 0.20 & 0.27 \\
        show-o & 0.28 & 0.36 & 0.40 & 0.23 & 0.33 & 0.22 & 0.30 \\
        vila-u-7b-256 & 0.26 & 0.33 & 0.37 & 0.35 & 0.39 & 0.23 & 0.31 \\
        show-o-512 & 0.28 & 0.40 & 0.48 & 0.30 & 0.46 & 0.30 & 0.35 \\
        Janus-Pro-7B & 0.30 & 0.37 & 0.49 & 0.36 & 0.42 & 0.26 & 0.35 \\
        Emu3 & 0.34 & 0.45 & 0.48 & 0.41 & 0.45 & 0.27 & 0.39 \\
        Liquid & 0.38 & 0.42 & 0.53 & 0.36 & 0.47 & 0.30 & 0.41 \\
        Harmon-1.5B & 0.38 & 0.48 & 0.52 & 0.37 & 0.44 & 0.29 & 0.41 \\
        OpenUni-B-512 & 0.37 & 0.45 & 0.58 & 0.39 & 0.50 & 0.30 & 0.43 \\
        Manzano-3B & 0.42 & 0.51 & 0.59 & 0.45 & 0.51 & 0.32 & 0.46 \\
        OpenUni-L-512 & 0.51 & 0.49 & 0.64 & 0.48 & 0.63 & 0.35 & 0.52 \\
        OpenUni-L-1024 & 0.49 & 0.53 & 0.69 & 0.49 & 0.56 & 0.39 & 0.52 \\
        BAGEL & 0.44 & 0.55 & 0.68 & 0.44 & 0.60 & 0.39 & 0.52 \\
        Manzano-30B & 0.58 & 0.50 & 0.65 & 0.50 & 0.55 & 0.32 & 0.54 \\
        MetaQuery-XL & 0.56 & 0.55 & 0.62 & 0.49 & 0.63 & 0.41 & 0.55 \\
        UniWorld-V1 & 0.53 & 0.55 & 0.73 & 0.45 & 0.59 & 0.41 & 0.55 \\
        Hunyuan-Image 3.0 & 0.58 & 0.57 & 0.70 & 0.56 & 0.63 & 0.31 & 0.57 \\
        UniWorld-V2 & 0.60 & 0.61 & 0.70 & 0.53 & 0.64 & 0.32 & 0.58 \\
        NextFlow-RL & 0.63 & 0.63 & 0.77 & 0.58 & 0.67 & 0.39 & 0.62 \\
        Qwen-Image & 0.62 & 0.63 & 0.77 & 0.57 & 0.75 & 0.40 & 0.62 \\
        LongCat-Image & 0.66 & 0.61 & 0.72 & 0.66 & 0.72 & 0.49 & 0.65 \\
        DeepGen1.0 & 0.72 & \textbf{0.81} & 0.70 & 0.67 & 0.82 & 0.66 & 0.73 \\
        GPT-4o & 0.81 & 0.71 & \textbf{0.89} & 0.83 & 0.79 & 0.74 & 0.80 \\
        
        \midrule
        
        \multicolumn{8}{c}{\textbf{\textit{Visual Generation Agent}}} \\ [0.5ex]
        
        GenAgent & 0.78 & 0.67 & 0.78 & 0.72 & 0.77 & 0.55 & 0.72 \\
        Gen-Searcher-8B + Qwen-Image & 0.80 & 0.71 & 0.82 & 0.76 & 0.74 & 0.75 & 0.77 \\
        Mind-Brush & 0.83 & 0.69 & 0.84 & 0.71 & \textbf{0.85} & 0.68 & 0.78 \\
        GenEvolve & 0.84 & 0.74 & 0.87 & 0.83 & 0.81 & \textbf{0.83} & 0.82 \\
        RS-Gen & \textbf{0.86} & 0.74 & 0.88 & \textbf{0.85} & 0.83 & 0.74 & \textbf{0.83} \\

        \bottomrule 
    \end{tabular*}
\end{table}

Considering that most existing methods conducted their experiments on the WISE\_legacy~\cite{niu2025wise} benchmark (the legacy version of WISE), we also perform evaluations on the WISE\_legacy~\cite{niu2025wise} benchmark to establish a more comprehensive comparison. Consistent with the previous experiments, the RS-Gen framework adopts Qwen-Image~\cite{wu2025qwenimagetechnicalreport} as the baseline foundational image generation model. In accordance with the official evaluation protocol of WISE\_legacy~\cite{niu2025wise}, gpt-4o-2024-05-13 is utilized as the evaluation model, and the experimental results are shown in Table \ref{tab:appendix_wise_legacy_result}. Compared to the baseline Qwen-Image~\cite{wu2025qwenimagetechnicalreport} (0.62), the RS-Gen framework using Qwen-Image~\cite{wu2025qwenimagetechnicalreport} as the underlying foundational generator demonstrates a remarkably powerful performance enhancement, achieving the highest comprehensive score (Overall: 0.83) among all solutions, and even outperforming the closed-source commercial model GPT-4o~\cite{openai2025gpt4o} (0.80). Specifically, within the visual generation agent paradigm solutions, RS-Gen leads previous state-of-the-art frameworks (such as GenEvolve~\cite{chen2026genevolve} and Mind-Brush~\cite{he2026mindbrush}) by an absolute advantage. Notably, on the Cultural and Biology subsets, RS-Gen reaches scores of 0.86 and 0.85, respectively. This significant performance gain is primarily attributed to our designed "propose-and-solve" scheme and dynamic error-correction mechanism. By dynamically identifying knowledge gaps and proactively retrieving external factual knowledge, RS-Gen successfully shatters the static knowledge cutoff limitations of the foundational model Qwen-Image, significantly suppressing the emergence of factual hallucinations when dealing with out-of-distribution (OOD) concepts or long-tail entities.

Furthermore, compared with traditional models, RS-Gen provides a plug-and-play, training-free solution for the open-source community. Although certain specialized models or commercial closed-source models may maintain a lead on a specific subset due to their massive parameter scales or domain-specific data tuning, RS-Gen achieves the most robust and balanced performance gains across almost all subsets without updating any model parameters. These experimental results fully demonstrate that by reconstructing the traditional "black-box mapping" image generation process into a multi-stage agentic collaborative workflow, RS-Gen can effectively resolve implicit user intentions, bridge knowledge gaps, and crack complex reasoning puzzles, firmly proving its exceptional generalization capability and superior performance.

\end{document}